\documentclass[pdflatex,sn-mathphys-num]{sn-jnl}

\usepackage{graphicx}%
\usepackage{multirow}%
\usepackage{amsmath,amssymb,amsfonts}%
\usepackage{amsthm}%
\usepackage{mathrsfs}%
\usepackage[title]{appendix}%
\usepackage{xcolor}%
\usepackage{textcomp}%
\usepackage{manyfoot}%
\usepackage{booktabs}%
\usepackage{algorithm}%
\usepackage{algorithmicx}%
\usepackage{algpseudocode}%
\usepackage{listings}%
\lstdefinelanguage{text}{}
\usepackage{siunitx}
\usepackage{tabularx}
\usepackage{array}
\usepackage{makecell}
\usepackage{caption}
\usepackage{diagbox}
\newcolumntype{L}[1]{>{\raggedright\arraybackslash}p{#1}}
\newcolumntype{C}[1]{>{\centering\arraybackslash}p{#1}}
\newcolumntype{Y}{>{\centering\arraybackslash}X}

\theoremstyle{thmstyleone}%
\theoremstyle{thmstyletwo}%

\theoremstyle{thmstylethree}%

\begin{document}

\title[KG-enhanced LLM for Incremental Game PlayTesting]{Knowledge Graph-enhanced Large Language Model for Incremental Game PlayTesting}


\author[1]{\fnm{Enhong} \sur{Mu}}
\author[2]{\fnm{Jinyu} \sur{Cai}}
\author[2]{\fnm{Yijun} \sur{Lu}}
\author[1]{\fnm{Mingyue} \sur{Zhang}}
\author[3]{\fnm{Kenji} \sur{Tei}}
\author*[2]{\fnm{Jialong} \sur{Li}}\email{lijialong@fuji.waseda.jp}

\affil[1]{\orgname{Southwest University}, \orgaddress{\country{China}}}
\affil[2]{\orgname{Waseda University}, \orgaddress{\country{Japan}}}
\affil[3]{\orgname{Institute of Science Tokyo}, \orgaddress{\country{Japan}}}


\abstract{The rapid iteration and frequent updates of modern video games pose significant challenges to the efficiency and specificity of testing. Although automated playtesting methods based on Large Language Models (LLMs) have shown promise, they often lack structured knowledge accumulation mechanisms, making it difficult to conduct precise and efficient testing tailored for incremental game updates. To address this challenge, this paper proposes a KLPEG framework. The framework constructs and maintains a Knowledge Graph (KG) to systematically model game elements, task dependencies, and causal relationships, enabling knowledge accumulation and reuse across versions. Building on this foundation, the framework utilizes LLMs to parse natural language update logs, identify the scope of impact through multi-hop reasoning on the KG, enabling the generation of update-tailored test cases. Experiments in two representative game environments, Overcooked and Minecraft, demonstrate that KLPEG can more accurately locate functionalities affected by updates and complete tests in fewer steps, significantly improving both playtesting effectiveness and efficiency.}




\maketitle

\section{Introduction}
\label{sec:intro}
The video game industry has grown into a multi-billion-dollar global market, with increasingly complex games demanding rigorous quality assurance processes to meet player expectations and maintain brand reputation. Game testing plays a critical role in ensuring functionality, stability, and user satisfaction, especially as modern games often involve large open worlds, real-time multiplayer interactions, and the continuous update cycles typical of live service games~\cite{Politowski2021A}. Effective testing not only helps uncover gameplay and interface issues but also prevents post-release regressions that can lead to negative user reviews and financial losses~\cite{ruuska2015quality}.

In practice, game testing has traditionally relied on manual playtesting or the handcrafting of test cases, where testers interact directly with the game to identify issues or verify specific behaviors. Although manual testing provides valuable qualitative feedback and adapts flexibly to diverse scenarios, it suffers from significant limitations such as high labor costs, inconsistency, and difficulty scaling to frequent and extensive game updates~\cite{ostrowski2013automated}.

To address the limitations of manual testing, automated playtesting methods have been widely explored. 
Early automated playtesting used scripted approaches; \cite{ostrowski2013automated} developed a model combining record-and-playback with scripting to support low-code regression testing~.
Recently, reinforcement learning (RL)-based approaches have emerged as a promising automated approach, in which RL agents dynamically interact with the game environment, autonomously uncovering edge cases and scenarios often overlooked by human testers~\cite{9619048}.  Hybrid approaches that combine genetic algorithms (GA) and deep RL further enhance exploration efficiency and bug detection by evolving more effective testing strategies over time~\cite{zhao2019wuji}. However, despite these advances, RL methods often require extensive retraining and parameter tuning whenever the game is updated, resulting in high maintenance costs and limited responsiveness to rapid iteration cycles.

More recently, natural language processing (NLP) techniques and large language models (LLMs) have been introduced into automated game testing. With strong capabilities in natural language understanding and reasoning, LLMs can comprehend game tasks, parse rules, and generate test plans. For example, \cite{taesiri2022large} demonstrated the potential of LLMs in zero-shot bug detection by formulating the task as question answering over textual descriptions of game events. 
\cite{paduraru2024unit} explored the use of LLMs to generate unit tests from natural language descriptions of game behavior, thereby improving testing efficiency and reducing manual effort. \cite{hu2024language} proposed a framework that combines online planning with NLP to accelerate the smoke testing of MMORPG quests, significantly reducing the time required for core functionality verification.

However, although LLM-based approaches are promising, existing studies have largely overlooked the practical realities of agile game development environments. Specifically, modern games often require frequent updates to continuously engage players by introducing new content. For instance, NetEase reportedly conducts three scheduled version iterations per day within its internal game development process~\cite{yuechen2020regression}. This real-world demand highlights two major limitations of current LLM-based methods: 
(1) Lack of structured knowledge accumulation and reuse mechanisms. Existing LLM-based testing methods rely heavily on zero-shot inference or multi-turn contextual generation without a unified knowledge representation. As a result, relationships among game elements, rules, and behaviors are difficult to persistently accumulate or reuse across similar scenarios, thereby limiting long-term effectiveness and consistency;  
(2) Difficulty in targeted testing tailored to the version updates. Current approaches generally perform holistic LLM reasoning over the game, lacking the structured understanding needed to isolate update-specific changes. Consequently, they struggle to support incremental testing or targeted validation of update-induced content.

To this end, this paper proposes \underline{K}nowledge Graph-Enhanced \underline{L}arge Language Model for \underline{P}layTesting \underline{E}volving \underline{G}ame (KLPEG), a Knowledge Graph(KG)-centered framework for LLM-driven automated game testing. The core idea is to leverage a KG that (1) captures essential dependencies among game elements and tasks, supporting cross-version knowledge retention; (2) uses LLMs to semantically parse natural language update logs and perform graph-based reasoning to identify affected game elements; and (3) enables the generation of targeted test cases for the game updates by guiding LLMs based on the localized impact inferred from the KG.

The rest of this paper is organized as follows. Section~\ref{sec:relatedwork} reviews related work in the related field. Section~\ref{sec:methodology} presents KLPEG in detail. Section~\ref{sec:evaluation} describes our experimental setup and discusses the results. Finally, Section~\ref{sec:conclusion} concludes the paper and outlines future research directions.

\section{Background and Related Work}
\label{sec:relatedwork}
\subsection{Large Language Models for Game Agents}
In recent years, LLMs have demonstrated strong capabilities in natural language understanding, action planning, and multi-step reasoning, making them widely adopted for developing intelligent game agents~\cite{10.1145/3686803,xu2024largelanguagemodelssynergize}. For instance, \cite{bateni2024language} proposed a zero-shot language agent that plays games like \textit{Slay the Spire} by generating action plans using natural language and chain-of-thought (CoT) reasoning, showing advantages in rule comprehension and long-term strategy planning.
Building on this, the Voyager system~\cite{wang2024voyager} advanced the field by using LLMs to generate reusable "skill functions" for solving game tasks, organizing them into a skill library to support continual learning. 
Following that, Cradle~\cite{tan2024towards}, a multimodal agent designed to complete complex tasks using screen images (and optionally audio) as input and keyboard/mouse actions as output, demonstrated strong generalization in games like \textit{Red Dead Redemption II}, \textit{Cities: Skylines}, and \textit{Stardew Valley}, all without API access.

Despite these advances in task completion and intelligent behavior generation, current approaches still fall short of meeting the practical needs of modern game testing. On one hand, they generally lack behavior traceability, version-change awareness, and structured modeling of task dependencies—features that are crucial for systematic, explainable, and version-sensitive validation workflows. On the other hand, testing is inherently a cost-sensitive phase of game development. Some methods—such as Cradle, which relies on high-end multimodal models like GPT-4V—incur high inference costs, further limiting their practicality and scalability in real-world testing environments.

\subsection{Game Testing Techniques}
Recent research in game testing presents diverse methodological trends. Some works aim to improve the systematicity and maintainability of testing through structured representations of behavior paths and state dependencies. For example, \cite{ostrowski2013automated} proposed a low-code framework based on script recording and playback, allowing testers to convert interactions into reproducible regression test cases and reduce redundant manual effort. However, this approach still depends on expert knowledge to identify key interactions. GameRTS~\cite{GameRTS} constructs a game state transition graph via static code analysis and selects affected test cases through version differencing, thereby enhancing the precision and efficiency of regression testing. Yet, this method assumes access to source code and relies on a white-box environment, which is often unavailable in real-world game development~\cite{Politowski2021A}.

Other works in playtesting emphasize active exploration of the game state space in the absence of prior knowledge, aiming to improve coverage and discover unexpected behaviors. To overcome the sparse reward problem in traditional RL, \cite{9619048} introduced curiosity-driven intrinsic rewards to guide agents toward less-visited states and improve behavioral diversity. Wuji~\cite{zhao2019wuji} further combined deep reinforcement learning with evolutionary strategies to evolve agent policies, striking a balance between task completion and exploration. Although to varying degrees, these methods mainly focus on exploration rather than targeted testing. While effective at uncovering hidden defects or anomalies, they often lack the ability to perform focused validation based on specific updates. Furthermore, the high retraining cost across versions limits their applicability in iterative development settings.

With the rapid advancement of LLMs in semantic understanding and reasoning, recent studies have explored their integration into game testing pipelines. These approaches leverage the ability of LLMs to interpret rules, tasks, and player behavior from natural language, lowering the barrier to testing automation. \cite{taesiri2022large} formulated bug detection as a question-answering task, enabling zero-shot reasoning over textual descriptions of game events. \cite{paduraru2024unit} used LLMs to generate unit test code from functional descriptions, reducing the burden of manual test design. LAGET~\cite{hu2024language} combined natural language task descriptions with optimal trajectory samples to train a language-guided testing agent, improving exploration efficiency and task success rates. These methods demonstrate strong adaptability in handling complex task descriptions and unstructured inputs such as update logs and design documents. However, they rely heavily on prompt engineering and contextual inference, and lack a unified and structured knowledge model, as well as long-term memory. As a result, they struggle to track semantic task dependencies or transfer testing experience across versions, limiting their stability and effectiveness in complex, update-driven testing scenarios.



\section{Knowledge Graph-Enhanced LLM PlayTesting}
\label{sec:methodology}
\begin{figure*}[htb]
\centering
\includegraphics[width=1.0\linewidth]{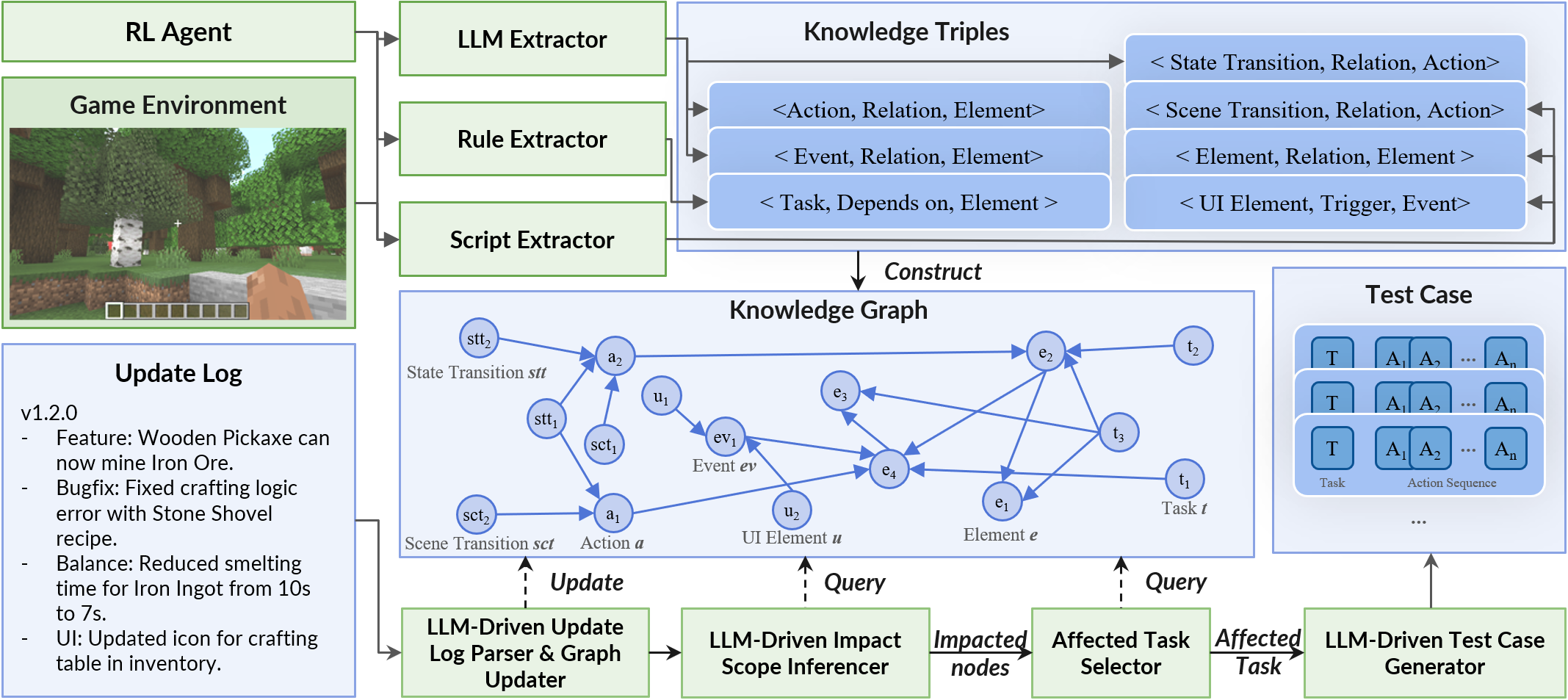}
\caption{KLPEG Overview. Data and module with blue and green backgrounds, respectively.
}
\label{fig:framework}
\end{figure*}

This section introduces KLPEG using a running example of Minecraft.
Briefly summarizing, KLPEG consists of three main stages.  
The first stage is game environment exploration and data collection, where an RL agent interacts with the game environment to collect comprehensive action–state trajectories that capture both typical and edge-case behaviors.  
The second stage is knowledge triple extraction and KG construction, where we leverage LLM-based semantic parsing in combination with game-specific scripts and rules to extract structured knowledge triples from the collected interaction traces.  
The third stage is update logs analysis, impact inference, and test case generation. In this stage, LLMs parse the natural language update logs to identify modified elements and relationships. Through multi-hop reasoning over the constructed KG, the framework infers the likely impact region of each update and automatically generates targeted test cases to validate those regions.  

\subsection{Game Environment Execution and Data Collection}
\label{sec:step1}
To provide a foundation of actionable and reusable knowledge, the first stage employs a curiosity-driven RL agent\cite{9619048} (using the PPO algorithm) to interact with the game environment, aiming to balance exploration and task completion for comprehensive coverage. This approach combines intrinsic and extrinsic rewards, where the intrinsic reward is novelty-based, encouraging exploration of rarely visited states through prediction error signals, while extrinsic rewards guide the agent toward task-relevant behaviors. This hybrid design ensures that the collected action-observation trajectories reflect both common and edge-case interactions.

\subsection{Knowledge Triple Extraction and Knowledge Graph Construction}
\label{sec:step2}

We use extractors based on LLMs, scripts, and rules to transform the game execution logs (from API) into structured, machine-interpretable knowledge.  
The core data structure of the knowledge is the knowledge triple, defined as \textit{(Node, Relation, Node)}. These triples are used to dynamically construct a KG $G = (V, E, R)$, where $V$ is the set of game elements, $R$ is the set of relation types, and $E$ is the set of directed, labeled edges. Each triple encodes a specific dependency or causal relationship. 

The LLM extractor leverages the semantic understanding capabilities of LLMs. It takes natural language as input and outputs structured triples. During gameplay, natural language observations and prompts are present and serve as inputs to the LLM-based extractor, which converts the raw data into knowledge triples via semantic parsing. For instance, after building a Nether portal, the game might prompt the player with “Touching the Nether portal will teleport you to the Nether.” Feeding this prompt into the LLM extractor yields the triple (Overworld, Touching Nether Portal, Nether).

The script-based extractor uses predefined scripts that take raw actions, state observations, and state comparisons to extract knowledge triples through several processing steps.
For instance, given a game execution log “a wooden pickaxe interacts with stone,” a script determines whether this interaction is valid and, if so, extracts the triple (Wooden Pickaxe, mines, Stone). 
To adapt knowledge extraction tasks to different games, the hand-crafted script extractor is implemented as a single Python module. It defines several exportable function interfaces (e.g., \texttt{on\_event(log\_line)} and \texttt{on\_state\_change(prev\_state, curr\_state)}), which are automatically called by the framework at runtime. For ease of development, a unified Python template is provided for extractor development.

In a rule-based extractor, by predefining rules (e.g., regular expressions), relevant triples can be extracted.

For example, given a log “the player uses wooden planks to craft stick”, the extractor would generate the triple (\textit{Wooden Plank}, crafts, \textit{Stick}).
Additonally, logs for crafting tasks like ``Craft Wooden Pickaxe,'' ``Craft Iron Pickaxe,'' and ``Craft Iron Sword'' all follow the pattern ``Craft [Item],'' so a regular expression such as \texttt{Craft\textbackslash s*(\textbackslash S+)} can be used to extract triples like (Craft Wooden Pickaxe, depends on, Wooden Pickaxe). 
The rule-based extractor is implemented as a JSON configuration file. For each extraction source (such as logs and task names), there is an array corresponding to it, and multiple regular expressions can be configured.

Note that the design of the KG needs to be customized and tailored to the characteristics of the game under test. Different game genres emphasize different types of relationships: strategy games often require modeling of resource dependencies and technological progression; action games need to model state transitions and causal links between actions and environmental changes; RTS games focus on unit interactions, build orders, and tactical confrontations; while card games require precise modeling of conditional triggers, card synergies, and turn-based effects. By aligning the knowledge representation with genre-specific mechanics and testing goals, the resulting KG can better support accurate reasoning and targeted test generation.

\begin{table}[h]
\centering
\caption{Main Types of Knowledge Triples.}
\begin{tabularx}{\textwidth}{@{}L{0.23\textwidth} L{0.35\textwidth} X@{}}
\toprule
\textbf{Type} & \textbf{Description} & \textbf{Triple Example (Head Ent., Relation, Tail Ent.)}\\
\midrule
Game Element Interaction & Relations between two game elements & (Wooden Pickaxe, mines, Stone) \\
Task Dependency & Tasks or goals dependent on game elements & (Obtain Stone Sword, depends\_on, Wooden Plank) \\
Causal Transition & Actions that cause state changes & (Raw Porkchop, cooked\_to, Cooked Porkchop) \\
UI-Action Mapping & UI elements triggering in-game events & (Clicking Crafting Result, triggers, Acquiring Iron Sword) \\
Scene Transition & Actions triggering scene transitions & (Overworld, transitions\_to, Nether) \\
Event-Element Relation & Relations between events and game elements & (Acquiring Iron Sword, associated with, Iron Sword) \\
Action-Element Relation & Relations between actions and game elements & (Smelting, associated with, Furnace) \\
\bottomrule
\end{tabularx}
\label{tab:triple_types}
\end{table}



\subsection{Update Log Analysis, Impact Scope Inference, and Test Case Generation}
\label{sec:step3}
This phase employs a four-stage pipeline that integrates update log analysis with structured reasoning over the KG.
The first step is to update log parsing and graph synchronization. LLMs extract changed elements, relations, or actions from natural language update logs. For example, given a log entry like “Wooden Pickaxe can now mine Iron Ore,” the framework extracts the new triple (\textit{Wooden Pickaxe}, \textit{mines}, \textit{Iron Ore}) and inserts it into the KG via symbolic operations such as \texttt{MinecraftGraphUpdate(op=insert, triple)}.

Following graph synchronization, the framework infers the scope of impacted content by performing multi-hop reasoning on the updated KG. In this example, the addition of a new mining capability affects elements connected to Iron Ore, such as crafting Iron Ingots or building Iron Tools. Formally, for each updated node $u$, the set of potentially impacted elements, denoted $\mathcal{I}_u$, is computed via bounded multi-hop traversal over the graph:

\begin{align}
\mathcal{I}_u = \left\{ v\ \middle|\ 
\begin{array}{l}
\exists\ \text{path } (u, r_1, v_1), (v_1, r_2, v_2), \ldots, \\
(v_{k-1}, r_k, v)\ \text{ in } G,\quad k \leq K
\end{array}
\right\}
\end{align}

where $K$ is the maximum hop threshold and $(a, r, b)$ denotes a directed edge labeled $r$ from node $a$ to node $b$ in the KG $G$. This bounded traversal identifies both direct and indirect dependencies, enabling precise localization of potential impact regions.

Once the impact region is identified, the system selects tasks that interact with or depend on the updated elements. Specifically, it cross-references the affected elements with existing tasks in the KG to identify those likely to be impacted, and generates a clear test objective. For example, if the original game includes a task like “craft an \textit{Iron Sword},” and the update enables the \textit{Wooden Pickaxe} to mine \textit{Iron Ore}, the generated test objective may be phrased as “verify the player can obtain \textit{Iron Ore} using a \textit{Wooden Pickaxe} and craft an \textit{Iron Sword}.”

The system then uses an LLM to perform reasoning over the knowledge graph and generate a detailed test case. In the context of playtesting, a test case generally refers to a testing objective (i.e., completing a specific task) and its corresponding operational steps (i.e., action sequence). For instance, given a test case whose task is ``making an Iron Sword'', the action sequence may include the following steps: collect \textit{wood}, craft a \textit{Wooden Pickaxe}, mine \textit{Iron Ore}, smelt it into \textit{Iron Ingots}, and finally combine \textit{Iron Ingots} with \textit{sticks} to craft an \textit{Iron Sword}.


\section{Evaluation}
\label{sec:evaluation}
The evaluation aims to answer the following research questions:
\begin{itemize}
\item \textbf{RQ1 (Effectiveness)}: To what extent can KLPEG effectively target and validate game functionalities impacted by updates, compared to baseline testing methods?
\item \textbf{RQ2 (Efficiency)}: How does KLPEG improve the overall efficiency of the testing workflow, particularly in terms of execution time and the number of steps required to validate an update?
\end{itemize}

\subsection{Experiment Settings}

\begin{figure}[h!tb]
    \centering
    \includegraphics[width=0.9\linewidth]{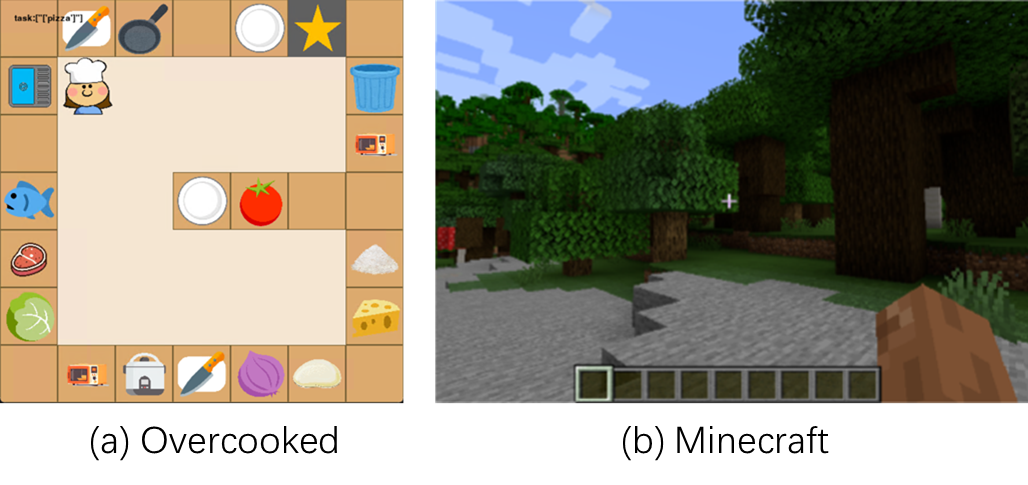}
    \caption{Screenshots of the Game Environment.}
    \label{fig:gameimage}
\end{figure}

\textbf{Target Games}.
We perform experiments in two well-known games: Overcooked\cite{overcooked2016} and Minecraft\cite{minecraft2009} (Fig.~\ref{fig:gameimage}). Overcooked is a 2D simulation game focused on completing cooking tasks, where players take on the role of a chef, interacting with vegetables and tools to fulfill various recipe tasks. Minecraft is an open-world 3D sandbox game where players can freely explore, build, and destroy.

\textbf{Update Logs.} 
Consistent with prior studies on game testing in software engineering~\cite{ariyurek2019automated, GameRTS, yuechen2020regression}, the simulated updates are designed to test the core game mechanics and operational logic.
Specifically, they include three representative categories: 
(i) feature additions, which introduce new abilities or items (e.g., ``Wooden Pickaxe can now mine Iron Ore''), 
(ii) bug fixes, which correct unintended behaviors (e.g., ``fix: knife caused ingredient loss''), and 
(iii) rule or balance adjustments, which modifies existing parameters or crafting quantities (e.g., ``fixed incorrect wood block crafting count''). 
They represent the most frequent and incremental types of updates in live-service and sandbox games like Minecraft. 
A detailed exaple of update logs used in the experiments is provided in the Appendix.

\textbf{Metrics}.
For RQ1, we assess (1) targeted element coverage: the number of updated game elements accessed during testing, out of a fixed total; (2) targeted element interaction ratio: the proportion of total interactions that involve update-related elements; and (3) update-related bug detection ratio: the number of update-induced tasks successfully detected, out of a fixed total,    where bugs are defined based on a predefined set~\cite{ariyurek2019automated, feldmeier2023fully} derived from expert experience and correspond to specific state–action pairs or in-game states that may occur along the paths to completing game tasks. A bug is flagged when such a state–action pair is executed or the predefined buggy state is reached. These bug flags do not interrupt the execution of the experiment but are recorded for final statistical analysis. To align with the paper's focus on bugs introduced by new features and updates, all injected bugs are localized within or adjacent to updated components described in the update logs. For RQ2, we assess (1) total test time, which refers to the end-to-end duration from receiving the update log to completing all test executions; and (2) average test steps, which indicate the average number of environment interaction steps per test case.

\textbf{Baselines}. The baselines for comparison include:  
(1) RANDOM, which explores the environment by selecting actions at random; 
(2) GA, which evolves action sequences using genetic mutation and crossover; 
(3) curiosity-driven PPO (CD-PPO) \cite{9619048}, which employs RL agents with internal curiosity reward for discovering unseen game states to improve exploration coverage.
(4) naive LLM approach, in which the KG is removed from our KLPEG framework \cite{paduraru2024unit}, i.e., the LLM directly analyzes the update log to generate test cases.

\textbf{LLM Settings}.
We use four different LLMs: gpt-4.1, gpt-4o, Qwen3-plus, and DeepSeek v3. 

\textbf{Detailed Settings in Overcooked}.
The evaluation on Overcooked is using an open-source implementation \cite{cai2024overcooked}, which is a grid world with discrete actions (e.g., \textit{pick up tomato}, \textit{chop ingredient}, \textit{submit dish}). States include player position, held items, and the status of environment objects. Tasks are modeled after real Overcooked recipes, and the update logs simulate actual game update logs, including new features and fixed bugs~\cite{ariyurek2019automated, feldmeier2023fully}.
The experiment includes three updates and 14 known injected bugs. Each update includes update log items (e.g., “feat: added steak dish”; “fix: knife caused ingredient loss”). A total of seven distinct tasks are predefined. LLM-based methods select relevant tasks from the list, while baselines execute all tasks sequentially.
For the RANDOM method, each task is executed using a randomly sampled action sequence with a maximum length of 100 steps.
Rewards in Overcooked are: -0.1 per step, +10 for each processed ingredient, +200 for a completed dish, and -5 for incorrect submission. GA uses a population size of 50, a sequence length of 100, and 150 iterations. CD-PPO is trained with 200,000 steps and uses intrinsic rewards for exploration. Early stopping is triggered after 7 stagnant evaluation intervals. KLPEG performs one update reasoning, one graph reasoning, and ten action generations per update.
All methods are executed for 20 independent runs. 

\textbf{Detailed Settings in Minecraft}.  
The Minecraft environment is used to evaluate the scalability of the proposed approach in more complex, open-world scenarios. Players start from a fixed position and can perform parameterized high-level commands such as \textit{mine}, \textit{craft}, and \textit{attack}, which are automatically decomposed into sequences of atomic operations. The environment state includes the player's location, inventory, and the dynamic status of spatial elements such as resource blocks, furnaces, and enemies.
Task design is based on Minecraft's built-in achievement system and includes nine tasks across three difficulty levels: basic collection, intermediate crafting, and advanced combat. The update setup also includes three simulated updates with realistic update log entries, including feature additions (e.g., “add diamond pickaxe”) and bug fixes (e.g., “fix issue with incorrect wood block crafting quantity”).
Rewards in Minecraft environment are: -0.1 for each step, +10 for each key component obtained, and +200 for each task completed.
A total of 15 update-related bugs are injected for evaluation.
The settings of each method are generally consistent with those used in the Overcooked environment, except that the maximum step limit for the RANDOM method is increased to 500 to accommodate the complexity of the open-world setting.

\subsection{Experimental Results}
\begin{table*}[!t]
\centering
\small
\setlength{\tabcolsep}{5pt}
\renewcommand{\arraystretch}{1.2}
\caption{Evaluation Results in Overcooked.}
\label{tab:metrics-overcooked}

\begin{tabular}{@{}l l c c c c c@{}}
\toprule
\multicolumn{2}{c}{\textbf{Method}} &
\makecell[c]{\textbf{Element}\\\textbf{Coverage} $\uparrow$\\(max=10)} &
\makecell[c]{\textbf{Element}\\\textbf{Interaction} $\uparrow$\\(\% / Count)} &
\makecell[c]{\textbf{Bug}\\\textbf{Detection} $\uparrow$\\(max=14)} &
\makecell[c]{\textbf{Test}\\\textbf{Time} $\downarrow$\\(s)} &
\makecell[c]{\textbf{Avg.}\\\textbf{Steps} $\downarrow$} \\
\midrule
\multicolumn{2}{l}{RANDOM} & 10.0 & 30\% (28.81) & 12.04 & 2.33 & 96.53 \\
\multicolumn{2}{l}{GA}     & 10.0 & 43\% (13.82) & 12.04 & 330.78 & 32.46 \\
\multicolumn{2}{l}{CD-PPO} & 10.0 & 63\% (13.43) & 9.94 & 880.95 & 21.26 \\
\midrule
\multirow[c]{4}{*}{\makecell[l]{KLPEG\\w/o KG}} & GPT-4.1     & 10.0 & 62\% (5.98)  & 13.02 & 34.42 & 9.62  \\
 & GPT-4o      & 10.0 & 50\% (4.79)  & 12.04 & 64.21 & 9.64  \\
 & Qwen3-Plus  & 10.0 & 57\% (5.43)  & 12.04 & 35.97 & 9.51  \\
 & DeepSeek V3 & 10.0 & 55\% (5.19)  & 12.04 & 208.19 & 9.49  \\
\cmidrule(l){2-7}
\multirow[c]{4}{*}{KLPEG} & GPT-4.1      & 10.0 & 90\% (9.16) & 14.00 & 40.70 & 10.13 \\
 & GPT-4o       & 10.0 & 90\% (8.72) & 14.00 & 71.05 & 9.64 \\
 & Qwen3-Plus   & 10.0 & 74\% (7.07) & 12.04 & 42.63 & 9.51 \\
 & DeepSeek V3  & 10.0 & 77\% (7.86) & 13.02 & 297.15 & 10.22 \\
\bottomrule
\end{tabular}

\end{table*}

Regarding the results of Overcooked, all methods achieved 100\% coverage of update-related elements, but they differed significantly in test focus and efficiency. KLPEG (GPT-4.1 / GPT-4o) demonstrated the best overall performance, achieving the highest interaction ratio (0.90), the lowest average number of steps (approximately 10), and a perfect bug detection ratio (1.00). In contrast, removing the KG module (w/o KG) resulted in a 20\%–40\% drop in interaction focus and reduced the bug detection ratio to 0.86–0.93. Among the baselines, RANDOM required the least time but performed poorly in both interaction focus and bug detection. GA exhibited moderate performance but incurred a high number of steps, indicating inefficiency in targeting update-sensitive areas. CD-PPO improved focus relative to RANDOM (interaction ratio 0.63) but suffered from high computational cost, requiring 21 steps and up to 880 seconds of total test time. 


\begin{table*}[!t]
\centering
\small
\setlength{\tabcolsep}{5pt}
\renewcommand{\arraystretch}{1.2}
\caption{Evaluation Results in Minecraft.}
\label{tab:metrics-mc}

\begin{tabular}{@{}l l c c c c c@{}}
\toprule
\multicolumn{2}{c}{\textbf{Method}} &
\makecell[c]{\textbf{Element}\\\textbf{Coverage} $\uparrow$\\(max=30)} &
\makecell[c]{\textbf{Element}\\\textbf{Interaction} $\uparrow$\\(\% / Count)} &
\makecell[c]{\textbf{Bug}\\\textbf{Detection} $\uparrow$\\(max=15)} &
\makecell[c]{\textbf{Test}\\\textbf{Time} $\downarrow$\\(s)} &
\makecell[c]{\textbf{Avg.}\\\textbf{Steps} $\downarrow$} \\
\midrule
\multicolumn{2}{l}{RANDOM} & 19.5 & 15\% (68.58) & 9.60 & 2.89 & 466.66 \\
\multicolumn{2}{l}{GA}     & 24.6 & 34\% (6.76)  & 10.65 & 227.29 & 19.94 \\
\multicolumn{2}{l}{CD-PPO} & 24.6 & 39\% (4.34)  & 10.65 & 749.84 & 11.01 \\
\midrule
\multirow[c]{4}{*}{\makecell[l]{KLPEG\\w/o KG}} & GPT-4.1     & 28.2 & 72\% (6.95)  & 13.95 & 29.91  & 9.62  \\
 & GPT-4o      & 27.9 & 59\% (5.72)  & 15.00 & 72.79  & 9.64  \\
 & Qwen3-Plus  & 27.9 & 61\% (5.82)  & 12.90 & 29.65  & 9.51  \\
 & DeepSeek V3 & 28.5 & 63\% (6.31)  & 13.95 & 217.67 & 10.00 \\
\cmidrule(l){2-7}
\multirow[c]{4}{*}{KLPEG} & GPT-4.1      & 30.0 & 92\% (6.01) & 15.00 & 35.91  & 6.50  \\
 & GPT-4o       & 30.0 & 91\% (7.73) & 15.00 & 75.03  & 8.52  \\
 & Qwen3-Plus   & 30.0 & 83\% (4.36) & 13.95 & 56.07  & 5.28  \\
 & DeepSeek V3  & 30.0 & 85\% (4.87) & 13.95 & 292.37 & 5.73  \\
\bottomrule
\end{tabular}

\end{table*}

For Minecraft, similar to the Overcooked environment, all LLM-based methods in Minecraft achieved full or near-complete coverage of update-related elements. Among them, KLPEG (GPT-4.1 / GPT-4o) again delivered the best performance, maintaining over 90\% interaction focus, averaging 6–8 test steps, and successfully detecting all 15 injected bugs. Notably, lightweight models such as Qwen3-Plus and DeepSeek V3 performed closer to the GPT-4 series than they did in Overcooked, detecting 14 bugs, increasing their interaction ratios to 83\%–85\%, and reducing average test steps to just 5–6—demonstrating a strong balance between accuracy and efficiency.
The w/o KG variants exhibited noticeable declines in both interaction focus and bug detection ratio, reaffirming the importance of structured graph-based reasoning for maintaining precision, especially in environments with complex logic.

Compared to Overcooked, the more complex experiments in Minecraft revealed more pronounced differences. Due to increased task complexity and spatial freedom, baseline methods deteriorated significantly: RANDOM’s coverage dropped from 100\% to 65\% (19.6 elements), with average steps exceeding 460. GA and CD-PPO retained moderate coverage (~82\%) but suffered from poor interaction ratios (34\%–39\%), failing to effectively converge on updated content. These results highlight the limitations of heuristic or exploratory approaches in sparse and highly dynamic environments, whereas structured LLM-based reasoning enables more targeted and efficient test generation.

\subsection{Discussion on Results}
\textbf{RQ1: Effectiveness}.
The experimental results demonstrate that KLPEG significantly outperforms baseline methods in terms of testing precision and coverage, with the advantage becoming more pronounced as game complexity increases. In the simpler Overcooked environment, all methods achieved full coverage of update-related elements, but KLPEG required substantially fewer steps and less time. In contrast, in the more complex and open-ended Minecraft environment, the performance of traditional methods deteriorated rapidly. For example, RANDOM and GA failed to fully cover the updated elements even after executing hundreds of actions, whereas KLPEG achieved 100\% coverage and detected all injected bugs with only a few steps. These findings indicate that the benefits of our approach are amplified as the state space and interaction freedom expand. By analyzing the scope of update impact and generating focused test cases, our method effectively avoids the inefficiency and randomness typical of conventional exploration-based approaches.

\textbf{RQ2: Efficiency}.
KLPEG offers significant efficiency advantages across the testing cycle. Once an update log is received, the entire process of reasoning and executing targeted tests completes within 40–70 seconds for Overcooked and 35–75 seconds for Minecraft—substantially faster than deep RL-based baselines like PPO, which require around 15 minutes per update due to retraining and exploration. In addition, KLPEG typically completes each test in just 6–10 steps, compared to dozens or even hundreds required by heuristic or random methods. A major contributor to this efficiency is KLPEG’s reusable KG, which maintains a persistent understanding of game mechanics and is incrementally updated, eliminating the need to retrain agents for each new version. These characteristics make KLPEG highly suitable for agile development settings that demand rapid, targeted testing.

\textbf{Effect of the Knowledge Graph}.
Ablation studies confirm the critical role of the KG module in enhancing test focus and bug detection rates. When the KG is removed, the proportion of interactions with updated elements drops by 20–30\%, and the detection ratio of update-related bugs declines significantly. This suggests that the KG provides a global structural view of game element dependencies, enabling LLMs to better understand relationships among elements and to filter out irrelevant paths. For example, in Overcooked, LLMs without access to the KG tend to repeatedly manipulate basic ingredients like tomatoes, while graph-guided agents are able to quickly focus on newly introduced tasks such as preparing steak-based dishes.

\textbf{Effect of LLM Foundation Performance}.
The results show that KLPEG is relatively robust to the choice of underlying LLM. The GPT-4 series consistently achieved the best performance across both environments, with over 90\% interaction focus and a 100\% bug detection ratio. Meanwhile, lighter models such as Qwen3-Plus and DeepSeek V3, when assisted by the KG, also produced competitive results—achieving interaction ratios of 83–85\% and bug detection ratios exceeding 90\%. These findings indicate that, with the KG, even mid-sized LLMs can support high-quality test generation with relatively strong effectiveness.

\subsection{Discussion and Limitations}
First, our approach relies on the completeness of the underlying KG extracted by the extractor. If the graph is missing critical mechanisms—such as hidden task branches—the update impact analysis may be incomplete. This highlights the practical challenge of constructing a comprehensive KG. As noted in prior work on KG construction and maintenance~\cite{peng2023knowledge}, potential strategies to address this limitation include behavior-based relation mining, automated graph completion using predictive models, and human-in-the-loop validation interfaces to ensure sufficient coverage of essential gameplay logic.

Second, in large-scale games such as open-world RPGs or MMORPGs, the number of graph nodes and relations may increase significantly, resulting in higher computational costs for querying and reasoning. Future research may explore strategies such as graph partitioning by module or scene, the use of graph databases for optimized querying, and the adoption of hierarchical graph structures to improve scalability.

Third, the framework's effectiveness is contingent upon the quality of the natural language update logs used to identify the scope of changes. To mitigate this inherent risk in our evaluation, the update logs were manually constructed to emulate the style and ambiguity of real-world logs. They intentionally preserved typical imperfections found in industry practice, such as (i) terse, task-focused language (e.g., `feat: added steak dish`), (ii) inconsistent verb usage (e.g., `fix`, `adjust`, `rework`), and (iii) the omission of causal details. This design reflects a realistic gray-box testing scenario where QA teams rely on such documentation. Nevertheless, this dependency remains a practical limitation; the system's ability to capture the true impact of an update would be compromised if faced with logs that are exceptionally vague, incomplete, or misleading. That said, it is worth noting that LLMs exhibit a certain degree of robustness when handling such ambiguity. Their semantic reasoning ability may allow them to infer plausible causal or structural relationships even when explicit information is missing. This capability helps partially mitigate the uncertainty and ambiguity introduced by incomplete update logs.

Fourth, a potential limitation is the framework's reliance on hand-crafted extractors, which could raise concerns about subjectivity and the reproducibility of knowledge extraction. However, in our approach, this manual process is highly constrained and objective. The function of the rule and script extractors is to apply deterministic transformation rules, designed by a developer, to convert game logs from the API into knowledge triples. Because this logic is directly mapped to the game's fixed log formats and the scope is limited to a predefined set of triple types, the implementation is consistent and reproducible regardless of who designs it. The extractors remain effective as long as the core log format is unchanged, but they would require a one-time manual update if the log format changes, such as during a major version update that introduces a new game system.

Fifth, another concern relates to the cost of developing custom extractors for each new game. In our study, this cost was minimal, requiring only about 30 minutes of development per game, even without reusing any templates. This efficiency stems from three key factors: (i) a narrowly focused extraction scope that targets only core gameplay logs while ignoring irrelevant entries (e.g., "User A has logged in"), (ii) the simplicity of the pattern-matching logic, which relies on a finite set of conditional statements (if-else) and regular expressions, and (iii) the well-structured nature of the logs in our tested environments. However, this low-cost model may not generalize to all scenarios. The development effort would likely increase significantly for game testing with more challenging data sources—such as obfuscated, unstructured, or purely visual logs—which would demand more advanced and costly techniques like causal reasoning or computer-vision-based template matching.

Sixth, our framework is deliberately designed to test a game's core mechanics and operational logic, a scope consistent with related studies in the software engineering field for game testing~\cite{ariyurek2019automated, GameRTS, yuechen2020regression}. The update types simulated in our experiments were chosen to reflect this focus, covering a diverse and realistic range of modifications to gameplay logic, including: (i) new functionality, (ii) bug fixes, and (iii) parameter or rule changes.
However, we acknowledge that this focused scope creates a boundary for our method's applicability. Many types of updates common in modern games fall outside what can be effectively modeled by a symbolic knowledge graph. These include changes to (i) user interface elements and layouts, (ii) non-functional aspects like 3D graphics rendering or physics engine behavior, and (iii) complex, data-driven systems such as game AI. Modeling and testing these areas often requires integrating specialized techniques from other domains, such as computer vision and physics simulation.

Finally, while KLPEG's focus on update-related regions allows it to excel in high-frequency, targeted testing scenarios, this targeted approach may overlook indirect regression bugs that emerge in seemingly stable, legacy parts of the game. To mitigate this limitation and extend the framework's applicability, future work could focus on two complementary strategies. First, KLPEG's generated test cases could serve as starting points for guided exploration. By subsequently employing complementary sampling strategies (e.g., multi-objective optimization), it could expand its coverage to a broader, yet still probabilistically update-relevant, state space. Second, KLPEG can be synergistically integrated with traditional regression testing; for instance, the dependency analysis capabilities of our Knowledge Graph could be leveraged to enhance the regression process itself by automatically prioritizing existing test cases from a legacy suite that are most likely to be affected by the changes, thus making large-scale regression testing more efficient.

\subsection{Threats to Validity}
For internal validity, first, the LLMs used in our framework are accessed via third-party APIs, and their pretraining processes are not under our control. These models may contain prior knowledge of the games or tasks, which could affect the neutrality of evaluation results.  
To mitigate this, we employed multiple LLMs from different vendors to reduce reliance on any single model.  
Second, threats may arise from baseline configuration and fault injection. RL and GA baselines may not have been trained with fully optimized hyperparameters, potentially underestimating their performance.  
To address these concerns, we followed the configuration practices in previous studies (e.g., sufficient training steps and population sizes).

For external validity, the first concern lies in the limited scope of evaluated games—only Overcooked and Minecraft were used.  
Although we chose two games with contrasting gameplay styles (2D simulation vs. 3D sandbox) to broaden scenario diversity, our method may not generalize to other genres, such as puzzle or FPS games.
Another threat concerns the realism of update logs. Our update logs were manually crafted to resemble real-world formats, but actual industry logs may be vague, incomplete, or intentionally obscured for security.  
While we attempted to reflect realistic update types (e.g., feature additions, bug fixes, balance changes), validating KLPEG under real development conditions will require future studies using production-grade update logs and version histories.

\section{Conclusion and Future Work}
\label{sec:conclusion}
This paper presents KLPEG, a KG-enhanced LLM playtesting framework tailored for game updates. Specifically, the KG is used both to model in-game elements, task dependencies, and causal relationships—enabling structured knowledge accumulation and cross-version reuse—and to support multi-hop reasoning with LLMs for identifying the impact scope of updates, thereby guiding the generation of targeted test cases. Experiments in two representative environments, Overcooked and Minecraft, demonstrate that KLPEG significantly outperforms traditional baselines across all key metrics.

Future work will focus on further enhancing the capability and applicability of KLPEG in two main directions:
First, we plan to enhance the scalability of KLPEG by adopting techniques such as graph partitioning, hierarchical representations, and optimized query mechanisms, enabling it to handle larger game environments with complex dependencies more effectively.
Second, we aim to expand the scope of KG modeling to encompass a broader range of elements, including UI logic, implicit gameplay rules, and environment-triggered events, thereby improving the system's ability to capture indirect update effects and support more comprehensive testing scenarios.

Third, we will investigate systematic approaches for generating and standardizing update logs to ensure testing efficiency and effectiveness. This includes developing automated log generation techniques that extract meaningful change descriptions from code commits and design documents, as well as establishing standardized formats for update logs that facilitate precise impact analysis within our KG framework.

Fourth, we plan to explore the application of KLPEG beyond game testing to other software domains. The framework's core capabilities, including structured knowledge representation, dependency analysis, and targeted test generation, are transferable to domains such as web applications and enterprise software that undergo frequent incremental updates.

\section*{Acknowledgement}
This work was conducted during the author's internship at Waseda University.
This research was supported by JSPS KAKENHI (No. 23K28064, 25K15290).

\appendix
\section*{Appendix: Prompt Design and Update Logs}

\lstset{
  backgroundcolor=\color{gray!5},
  frame=single,
  rulecolor=\color{gray!60},
  breaklines=true,
  breakatwhitespace=true,
  columns=flexible,
  keepspaces=true,
  aboveskip=0.5em,
  belowskip=0.5em,
  captionpos=b,
  basicstyle=\ttfamily\scriptsize,
}

This appendix provides representative examples of the LLM prompts and game update logs used in our experiments.

\begin{lstlisting}[language=text, caption={Prompt for LLM Extractor.}, label={lst:prompt_extractor}]
You are a game data extractor.
Your task is to extract game knowledge from the input natural language text.
Each piece of knowledge represents a relation between an action, event, or state transition and its associated object or result.

Requirements:
- Each knowledge item must contain three parts: Head, Relation, and Tail.
- The Head can be an action, event, or state transition.
- The Relation should accurately describe dependency, trigger, or causal connections between them.
- Keep the wording concise and do not provide explanations.
- If no extractable content exists, output an empty array.
- The output must be valid JSON in the following structure:

{
  "knowledge_triples": [
    {"Head": "...", "Relation": "...", "Tail": "..."},
    ...
  ]
}

Please extract knowledge from the following text: {input_text}

You may use the following relations: {relations}
\end{lstlisting}

\begin{lstlisting}[language=text, caption={Prompt for Update Log Parser \& Graph Updater.}, label={lst:prompt_parser}]
You are a game update log analyzer.
A set of graph-update tools has been provided to you.
You must explicitly invoke these tools during your reasoning process.

Your task:
1. Identify newly added or modified triples and involved elements from the update log.
2. Use the provided tools to expand or update the related nodes in the knowledge graph.
3. Output the final result in JSON format.
4. Do not include any explanations or additional text.

Output format:
{
  "new_or_modified_triples": [
    {"Head": "...", "Relation": "...", "Tail": "..."}
  ],
  "related_items": ["...",]
}

The update log is as follows: {update_log}
\end{lstlisting}

\begin{lstlisting}[language=text, caption={Prompt for Impact Scope Inferencer.}, label={lst:prompt_inferencer}]
You are a game impact scope inferencer.
A set of graph-reasoning tools has been provided to you.
You must explicitly use these tools during the inference process.

Your task:
1. Receive a given game element (item).
2. Invoke the provided reasoning tools to locate nodes in the knowledge graph that are directly or indirectly related to this element.
3. Output the inference results in JSON format.
4. Do not include any explanations or descriptive text.

Output format:
{
  "input_item": "...",
  "inferred_related_items": ["...",]
}

The input element is as follows: {input_item}
\end{lstlisting}

\begin{lstlisting}[language=text, caption={Prompt for Test Case Generator.}, label={lst:prompt_generator}]
You are a game automated test case generator.
Based on the input describing the affected content, generate a test case.

Requirements:
- Use the provided available actions to compose the action list.
- Do not include any explanations, comments, or additional text.
- The output must be valid JSON with the following structure:

{
  "Test_Objective": "...",
  "Action_Steps": ["...",]
}

Input: {impact_description}
\end{lstlisting}

\begin{lstlisting}[language=text, caption={Example of update log for Overcooked.}, label={lst:log_overcooked}]
# Game Update Log
## v1.2.0 -> v1.2.1

### Features
- Added new "Steak Dish" recipe.
- The chopping board now supports parallel ingredient cutting.

### Bug Fixes
- Fixed issue where the knife occasionally caused ingredient loss.
- Fixed serving window not updating after failed dish submission.

### Improvements
- Improved kitchen UI responsiveness and animation timing.
\end{lstlisting}

\begin{lstlisting}[language=text, caption={Example of update log for Minecraft.}, label={lst:log_minecraft}]
# Game Update Log
## v1.0.0 -> v1.0.1

### Features
- Introduced the Diamond Pickaxe, capable of mining all blocks at high speed.
- Introduced the Spider mob, which can be attacked using any sword.
- The Wooden Pickaxe now can mine Iron Ore.

### Bug Fixes
- Fixed a crash issue occurring when smelting sand in the furnace.

### Improvements
- Addressed various minor bugs and improved overall stability.
\end{lstlisting}

\bibliography{ref}

\end{document}